\documentclass{article}

\usepackage[T1]{fontenc}
\usepackage{tgtermes}
\usepackage[utf8]{inputenc}

\usepackage{amssymb}
\newcommand{\mathbold}[1]{\ensuremath{\boldsymbol{\mathbf{#1}}}}

\usepackage{times}

\usepackage{amsmath}
\usepackage{amsthm}

\newtheorem*{theorem*}{Theorem}
\newtheorem{proposition}{Proposition}

\usepackage[english]{babel}
\usepackage[parfill]{parskip}
\usepackage{afterpage}
\usepackage{enumitem}
\usepackage{framed}
\usepackage{xspace}

\renewenvironment{leftbar}[1][\hsize] {%
  \MakeFramed{\hsize#1\advance\hsize-\width\FrameRestore}%
}%
{\endMakeFramed}

\usepackage{lineno}

\usepackage{ragged2e}

\newcounter{parcount}

\DeclareRobustCommand{\parhead}[1]{\textbf{#1}~}
\usepackage[usenames,dvipsnames]{xcolor}
\definecolor{shadecolor}{gray}{0.9}

\usepackage{graphicx}
\usepackage[labelfont=bf]{caption}
\usepackage[format=hang]{subcaption}

\usepackage{booktabs,array}

\usepackage[algoruled]{algorithm2e}
\usepackage{listings}
\usepackage{fancyvrb}
\fvset{fontsize=\small}

\usepackage[numbers]{natbib}
\usepackage[colorlinks,linktoc=all]{hyperref}
\usepackage[all]{hypcap}
\hypersetup{citecolor=MidnightBlue}
\hypersetup{linkcolor=MidnightBlue}
\hypersetup{urlcolor=MidnightBlue}

\usepackage[nameinlink]{cleveref}
\crefname{section}{§}{§§}
\Crefname{section}{§}{§§}
\creflabelformat{equation}{#2\textup{#1}#3}  

\usepackage
[acronym,smallcaps,nowarn,section,nonumberlist]{glossaries}
\glsdisablehyper{}

\definecolor{strings}{rgb}{.624,.251,.259}
\definecolor{keywords}{rgb}{.224,.451,.686}
\definecolor{comment}{rgb}{.322,.451,.322}

\lstdefinelanguage{python}{
  morekeywords={from, import, as, for, in, while, def, return, =, +,
  -, /, *, lambda},
  keywords=[2]{build_toy_dataset, neural_network, __init__},
  keywords=[3]{Normal, Bernoulli, Beta, Binomial, Categorical, Dirichlet,
  Exponential, MultivariateNormalTriL, RandomVariable, Distribution,
  DirichletProcess, Empirical, PointMass, Gamma,
  MAP, Inference, KLqp, HMC, SGLD, SGHMC, KLpq,
  VariationalInference, MonteCarlo, ConjugateInference, GANInference,
  rnn_cell, dirichlet_process, cond, body,
  evaluate, ppc, copy, dot, get_session},
  morecomment=[l]{\#},
  morecomment=[s]{"""}{"""},
  morestring=[b]',
  morestring=[b]",
  alsoletter={<>=-+/*},
  sensitive=true
}

\usepackage{iclr2018_conference}
\newacronym{VI}{vi}{variational inference}
\newacronym{KL}{kl}{Kullback-Leibler}
\newacronym{ELBO}{elbo}{\emph{evidence lower bound}}
\newacronym{MCMC}{mcmc}{Markov chain Monte Carlo}
\newacronym{GWAS}{gwas}{genome-wide association studies}
\newacronym{SEM}{sem}{structural equation model}
\newacronym{SNP}{snp}{single nucleotide polymorphism}
\newacronym{LFVI}{lfvi}{likelihood-free variational inference}

\newcommand{\g}{\,|\,}

\renewcommand{\d}[1]{\ensuremath{\operatorname{d}\!{#1}}}

\newcommand{\mbb}{\mathbold{b}}

\newcommand{\mbv}{\mathbold{v}}
\newcommand{\mbw}{\mathbold{w}}
\newcommand{\mbx}{\mathbold{x}}
\newcommand{\mby}{\mathbold{y}}
\newcommand{\mbz}{\mathbold{z}}

\newcommand{\mbalpha}{\mathbold{\alpha}}

\usepackage{tikz}

\usetikzlibrary{bayesnet} 

\pgfdeclarelayer{edgelayer}
\pgfdeclarelayer{nodelayer}
\pgfsetlayers{edgelayer,nodelayer,main}

\definecolor{hexcolor0xbfbfbf}{rgb}{0.749,0.749,0.749}

\tikzset{>=latex}
\tikzstyle{none}   = [inner sep=0pt]
\tikzstyle{line}   = [ -, thick, shorten <=1pt, shorten >=1pt ]
\tikzstyle{arrow}  = [ ->, thick, shorten <=1pt, shorten >=1pt ]
\tikzstyle{ardash} = [ dashed, ->, thick, shorten <=1pt, shorten >=1pt ]

\tikzstyle{empty}=[circle,opacity=0.0,text opacity=1.0,inner sep=0pt]
\tikzstyle{box}=[rectangle,fill=White,draw=Black]
\tikzstyle{filled}=[circle,thick,fill=hexcolor0xbfbfbf,draw=Black]
\tikzstyle{hollow}=[circle,thick,fill=White,draw=Black]
\tikzstyle{param}=[rectangle,fill=Black,draw=Black,inner sep=0pt,minimum width=4pt,minimum height=4pt]
\tikzstyle{paramhollow}=[rectangle,thick,fill=White,draw=Black,inner sep=0pt,minimum
width=4pt,minimum height=4pt]

\usepackage{pgfplots}                               
\pgfplotsset{compat=newest}
\pgfplotsset{plot coordinates/math parser=false}
\newlength\figureheight
\newlength\figurewidth
\newlength\figureheightsmall
\newlength\figurewidthsmall
\definecolor{POSTcolor}{rgb}{0.48, 0.20, 0.58} 
\definecolor{Qcolor}{rgb}{0.00, 0.53, 0.22} 

\title{%
Implicit Causal Models for \\ Genome-wide Association Studies
}

\author{%
Dustin Tran \\
Columbia University \\
\And
David M. Blei \\
Columbia University \\
}

\iclrfinalcopy

\begin{document}

\maketitle

\begin{abstract}
Progress in probabilistic generative models has accelerated,
developing richer models with neural architectures, implicit densities,
and with scalable algorithms for their Bayesian inference.
However, there has been limited progress in
models that capture
causal relationships, for example, how individual genetic factors
cause major human diseases.
In this work, we focus on two challenges in particular: How do we build richer
causal models, which can capture highly nonlinear relationships and
interactions between multiple causes?
How do we adjust for latent confounders, which are variables
influencing both cause and effect and which
prevent learning of causal relationships?
To address these challenges, we synthesize ideas from causality and
modern probabilistic modeling. For the first, we describe
\emph{implicit causal models},
a class of causal models that leverages neural architectures
with an implicit density. For
the second, we describe an implicit causal model that adjusts for
confounders by sharing strength across examples.
In experiments, we scale Bayesian inference on up to a billion genetic
measurements. We achieve state of the art accuracy for identifying
causal factors: we significantly outperform existing genetics methods by
an absolute difference of 15-45.3\%.
\end{abstract}

\vspace{-1.5ex}
\section{Introduction}
\label{sec:introduction}
\vspace{-1ex}

Probabilistic models provide a language for specifying rich and
flexible generative processes
\citep{pearl1988probabilistic,murphy2012machine}.~Recent advances expand this language with neural architectures,
implicit densities, and with scalable algorithms for their Bayesian inference
\citep{rezende2014stochastic,tran2017deep}.~However, there has been limited progress in
models that capture high-dimensional causal relationships
\citep{pearl2000causality,spirtes1993causation,imbens2015causal}.
Unlike models which learn statistical relationships, causal models let
us manipulate the generative process and make counterfactual
statements, that is, what would have happened if the
distributions changed.

As the running example in this work, consider
\glsreset{GWAS}\gls{GWAS}
\citep{yu2005unified,price2006principal,kang2010variance}.  The goal
of \gls{GWAS} is to understand how genetic factors, i.e.,
\glspl{SNP}, cause traits to appear in individuals. Understanding this
causation both lets us predict whether an individual has a genetic
predisposition to a disease and also understand how to cure
the disease by targeting the individual \glspl{SNP} that cause it.

With this example in mind, we focus on two challenges to combining
modern probabilistic models and causality. The first is to develop
richer, more expressive causal models. Probabilistic causal models
represent variables as deterministic functions of noise and other
variables, and existing work usually focuses on additive noise models
\citep{hoyer2009nonlinear} such as linear mixed models
\citep{kang2010variance}.  These models apply simple nonlinearities
such as polynomials, hand-engineered low order interactions between
inputs, and assume additive interaction with Gaussian noise.  In
\gls{GWAS}, strong evidence suggests that susceptibility to common
diseases is influenced by epistasis (the interaction between multiple
genes) \citep{culverhouse2002perspective,mckinney2006machine}.  We
would like to capture and discover such interactions. This requires
models with nonlinear, learnable interactions among the inputs and the
noise.

The second challenge is how to address latent population-based
confounders.  In \gls{GWAS}, both latent population structure, i.e.,
subgroups in the population with ancestry differences, and relatedness
among sample individuals produce spurious correlations among
\glspl{SNP} to the trait of interest.  Existing methods correct for
this correlation in two stages
\citep{yu2005unified,price2006principal,kang2010variance}: first,
estimate the confounder given data; then, run standard causal
inferences given the estimated confounder.  These methods are
effective in some settings, but they are difficult to understand as
principled causal models, and they cannot easily accommodate complex
latent structure.

To address these challenges, we synthesize ideas from causality and
modern probabilistic modeling.  For the first challenge, we develop
\emph{implicit causal models}, a class of causal models that leverages
neural architectures with an implicit density.  With \gls{GWAS},
implicit causal models generalize previous methods to capture
important nonlinearities, such as gene-gene and gene-population
interaction.
Building on this, for the second challenge, we
describe an implicit causal model that adjusts for
population-confounders by sharing strength across examples (genes). We
derive conditions that prove the model consistently estimates the causal
relationship. This theoretically justifies existing methods and
generalizes them to more complex latent variable models of the confounder.

In experiments, we scale Bayesian inference on implicit causal models
on up to a billion genetic measurements.
Validating these results are not
possible for observational data \citep{pearl2000causality},
so we first perform an
extensive simulation study of 11 configurations of 100,000 \glspl{SNP} and 940 to 5,000 individuals.
We achieve state of the art accuracy for identifying
causal factors: we significantly outperform existing genetics methods by
an absolute difference of 15-45.3\%.
In a real-world \gls{GWAS}, we also show
our model discovers real causal relationships---identifying
similar \glspl{SNP} as previous state of the art---while being more
principled as a causal model.

\vspace{-1ex}
\subsection{Related work}
\label{sec:related}
\vspace{-1ex}

There has been growing work on richer causal models.
\citet{louizos2017causal}
develop variational auto-encoders for causality and address local
confounders via proxy variables. Our work is complementary: we develop
implicit models for causality and address global confounders
by sharing strength across examples.
In other work, \citet{mooij2010probabilistic} propose a
Gaussian process over causal mechanisms, and
\citet{zhang2009on} study post-nonlinear models,
which apply a nonlinearity after adding noise.
These models typically focus on the task of causal discovery, and they
assume fixed nonlinearities or smoothness which we
relax using neural networks.
In the potential outcomes literature,
much recent work has considered decision trees and neural networks
(e.g., \citet{hill2011bayesian,wager2015estimation,johansson2016learning}).
These methods tackle a related but different problem
of balancing covariates across treatments.

Causality with
population-confounders has primarily been studied for
\glsreset{GWAS}\gls{GWAS}.
A popular approach is to first, estimate the confounder using the top
principal components of the genotype matrix of
individuals-by-\glspl{SNP}; then, linearly regress the trait of
interest onto the genetic factors and these components
\citep{price2006principal,astle2009population}.  Another approach is
to first, estimate the confounder via a ``kinship matrix'' on the
genotypes; then, fit a linear mixed model of the trait given genetic
factors, and where the covariance of random effects is the kinship
matrix \citep{yu2005unified,kang2010variance}. Other work adjusts for
the confounder via admixture models and factor analysis
\citep{song2015testing,hao2016probabilistic}.  This paper builds on
all these methods, providing a theoretical understanding about when
causal inferences can succeed while adjusting for latent
confounders. We also develop a new causal model with nonlinear,
learnable gene-gene and gene-population interactions; and we describe
a Bayesian inference algorithm that
justifies the two-stage estimation.

The problem of epistasis, that is, nonadditive interactions
between multiple genes, dates back to classical work on epigenetics by
Bateson and R.A. Fisher
\citep{fisher1918correlation}.
Primary methods for capturing epistasis include adding
interactions within a linear model, permutation tests, exhaustive
search, and multifactor dimensionality reduction
\citep{cordell2009detecting}. These methods require hand-engineering
over all possible interactions, which grows exponentially
in the number of genetic factors.
Neural networks have been applied to address epistasis for
epigenomic data,
such as to predict sequence specificities of protein bindings
given DNA sequences \citep{alipanahi2015predicting}.
These methods use discriminative neural networks (unlike neural
networks within a generative model), and they focus on
prediction rather than causality.

\vspace{-1ex}
\section{Implicit Causal Models}
\label{sec:method}
\vspace{-1ex}

We describe the framework of probabilistic causal models. We then
describe implicit causal models, an extension of implicit
models for encoding complex, nonlinear causal relationships.

\subsection{Probabilistic Causal Models}

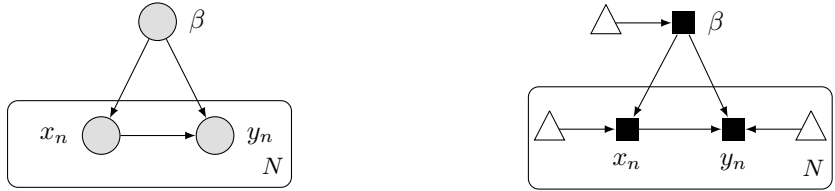
\begin{figure}[tb]
\begin{subfigure}{0.5\columnwidth}
  \centering
  \begin{tikzpicture}

  \node[obs, minimum size=0.5cm] (y)    {} ;
  \node[obs, left=of y, minimum size=0.5cm] (x)    {} ;

  \node[obs, above=of x, xshift=0.75cm, minimum size=0.5cm] (beta)    {} ;

  \node[right=0.04cm of beta] (betalabel) {$\beta$};
  \node[right=0.04cm of y] (ylabel) {$y_n$};
  \node[left=0.04cm of x] (xlabel) {$x_{n}$};

  \edge{x}{y};
  \edge{beta}{y};
  \edge{beta}{x};

  \plate[inner sep=0.2cm,xshift=-0.1cm,
    label={[xshift=-15pt,yshift=15pt]south east:$N$}] {plate1} {
    (y)(ylabel)(x)(xlabel)
  } {};

\end{tikzpicture}
\end{subfigure}
\begin{subfigure}{0.5\columnwidth}
  \centering
  \begin{tikzpicture}

  \factor[minimum size=0.3cm] {y} {} {} {};
  \factor[left=1.1cm of y, minimum size=0.3cm] {x} {} {} {};

  \factor[above=1.1cm of x, xshift=0.75cm, minimum size=0.3cm] {beta} {} {} {};

  \node (epsx) [minimum size=0.0cm,left=0.75cm of x] {};
  \node (epsxa) [minimum size=0.0cm,above=0.12cm of epsx.center] {};
  \node (epsxb) [minimum size=0.0cm,below=0.02cm of epsx.center, xshift=-0.2cm] {};
  \node (epsxc) [minimum size=0.0cm,below=0.02cm of epsx.center, xshift=0.2cm] {};
  \draw[] (epsxa.center)--(epsxb.center)--(epsxc.center)--(epsxa.center);

  \node (epsy) [minimum size=0.0cm,right=0.75cm of y] {};
  \node (epsya) [minimum size=0.0cm,above=0.12cm of epsy.center] {};
  \node (epsyb) [minimum size=0.0cm,below=0.02cm of epsy.center, xshift=-0.2cm] {};
  \node (epsyc) [minimum size=0.0cm,below=0.02cm of epsy.center, xshift=0.2cm] {};
  \draw[] (epsya.center)--(epsyb.center)--(epsyc.center)--(epsya.center);

  \node (epsbeta) [minimum size=0.0cm,left=0.75cm of beta] {};
  \node (epsbetaa) [minimum size=0.0cm,above=0.12cm of epsbeta.center] {};
  \node (epsbetab) [minimum size=0.0cm,below=0.02cm of epsbeta.center, xshift=-0.2cm] {};
  \node (epsbetac) [minimum size=0.0cm,below=0.02cm of epsbeta.center, xshift=0.2cm] {};
  \draw[] (epsbetaa.center)--(epsbetab.center)--(epsbetac.center)--(epsbetaa.center);

  \node[right=0.04cm of beta] (betalabel) {$\beta$};
  \node[below=0.04cm of y] (ylabel) {$y_n$};
  \node[below=0.04cm of x] (xlabel) {$x_{n}$};

  \edge{x}{y};
  \edge{beta}{x};
  \edge{beta}{y};

  \edge{epsx}{x};
  \edge{epsy}{y};
  \edge{epsbeta}{beta};

  \plate[inner sep=0.15cm,yshift=0.25cm,
    label={[xshift=-15pt,yshift=15pt]south east:$N$}] {plate1} {
    (y)(ylabel)(x)(xlabel)
    (epsy)
    (epsx)
  } {};

\end{tikzpicture}
\end{subfigure}%
\caption{Probabilistic causal model.
\textbf{(left)} Variable $x$ causes $y$ coupled with a shared variable $\beta$.
\textbf{(right)} A more explicit diagram where variables
(denoted with a square) are a deterministic function of other
variables and noise $\epsilon$ (denoted with a triangle).
}
\label{fig:causal_model}
\end{figure}

Probabilistic causal models \citep{pearl2000causality}, or structural
equation models,
represent variables as deterministic functions of noise and other variables.
As illustration,
consider the causal diagram in \Cref{fig:causal_model}.
It represents a causal model where there is a global variable
\begin{align*}
\beta &= f_\beta(\epsilon_{\beta}),\hspace{4em} \epsilon_{\beta}\sim s(\cdot),
\end{align*}
and for each data point $n=1,\ldots,N$,
\begin{align}
\begin{split}
x_n &= f_x(\epsilon_{x,n}, \beta),\hspace{5.55em} \epsilon_{x,n}\sim s(\cdot) \\
y_n &= f_y(\epsilon_{y,n}, x_n, \beta),\hspace{4em} \epsilon_{y,n}\sim s(\cdot).
\label{eq:causal_model}
\end{split}
\end{align}
The noise $\epsilon$ are background variables, representing unknown
external quantities which are jointly independent. Each
variable $\beta,x,y$ is a function of other variables and its background variable.

We are interested in estimating the causal mechanism $f_y$.
It lets us calculate quantities such as the causal effect
$p(y\g\operatorname{do}(X=x), \beta)$, the probability of an outcome
$y$ given that we force $X$ to a specific value $x$ and under fixed
global structure $\beta$.
This quantity differs from the conditional $p(y\g x, \beta)$. The
conditional takes the model and filters to the subpopulation where
$X=x$; in general, the processes which set $X$ to that value may also
have influenced $Y$. Thus the conditional is not the same as if we had
manipulated $X$ directly~\citep{pearl2000causality}.

Under the causal graph of \Cref{fig:causal_model},
the adjustment formula says that
$p(y\g\operatorname{do}(x),\beta)=p(y\g x,\beta)$.
This means we can estimate $f_y$ from observational data
$\{(x_n, y_n)\}$, assuming we observe the global structure
$\beta$.
For example, an additive noise model \citep{hoyer2009nonlinear} posits
\begin{equation*}
  y_n = f(x_n, \beta\mid\theta) + \epsilon_n,\qquad \epsilon\sim s(\cdot),
\end{equation*}
where $f(\cdot)$ might be a linear function of the concatenated inputs, $f(\cdot)=[x_n, \beta]^\top \theta$, or
it might use spline functions for nonlinearities.
If $s(\cdot)$ is standard normal,
the induced density for $y$ is normal with unit variance.
Placing a prior over parameters $p(\theta)$,
Bayesian inference yields
\begin{equation}
  \label{eq:posterior}
  p(\theta \g \mbx, \mby, \beta) \propto
  p(\theta) p(\mby \g \mbx, \theta, \beta).
\end{equation}
The right hand side is a joint density whose individual components can
be calculated. We can use standard algorithms, such as variational
inference or MCMC \citep{murphy2012machine}.

A limitation in additive noise models that they
typically apply simple
nonlinearities such as polynomials,
hand-engineered low-order interactions between inputs,
and assume additive interaction with Gaussian noise.
Next we describe how to build richer causal models which relax these
restrictions.
(An additional problem is that we typically don't observe $\beta$;
we address this in \Cref{sec:confounders}.)

\subsection{Implicit Causal Models}
\label{sub:implicit}

Implicit models capture an unknown distribution by hypothesizing about
its generative process \citep{diggle1984monte,tran2017deep}.
For a distribution $p(x)$ of observations $x$,
recent advances define a function $g$
that takes in noise $\epsilon \sim s(\cdot)$
and outputs $x$ given parameters $\theta$,
\begin{equation}
\label{eq:implicit}
x = g(\epsilon \g \theta),
\quad \epsilon\sim s(\cdot).
\end{equation}
Unlike models which assume additive noise, setting $g$ to be a neural
network enables multilayer, nonlinear interactions.
Implicit models also
separate randomness from the transformation;
this imitates the
structural invariance of causal models (\Cref{eq:causal_model}).

To enforce causality,
we define an \emph{implicit causal model} as a
probabilistic causal model where
the functions $g$ form structural equations,
that is, causal relations among variables.
Implicit causal models extend implicit models in the same way that
causal networks extend Bayesian networks
\citep{pearl1991theory} and path analysis extends regression analysis \citep{wright1921correlation}.
They are nonparametric structural equation
models where the functional forms are themselves learned.

A natural question is the representational capacity of implicit causal models.
In general, they are universal approximators:
we can use a fully connected network with a sufficiently large number of hidden
units to approximate each causal mechanism.
We describe this formally.

\begin{theorem*}[Universal Approximation Theorem]
Let the tuple $(\mathcal{E}, V, F, s(\mathcal{E}))$ denote a probabilistic causal model,
where $\mathcal{E}$ represent the set of background variables with
probability $s(\mathcal{E})$,
$V$ the set of endogenous variables, and $F$ the causal mechanisms.
Assume each causal mechanism is a continuous function on the
$m$-dimensional unit cube
$f\in \mathcal{C}([0,1]^{m})$.
Let $\sigma$
be a nonconstant, bounded, and monotonically-increasing continuous function.

For each causal mechanism $f$ and any error $\delta >0$,
there exist
parameters $\theta=\{\mbv,\mbw,\mbb\}$, for real constants
$v_{i},b_{i}\in {\mathbb  {R}}$ and real vectors
${\displaystyle w_{i}\in \mathbb {R} ^{m}}$ for
$i=1,\ldots,H$ and fixed $H$, such that the following function approximates $f$:
\begin{equation*}
g(x\g\theta)=\sum _{{i=1}}^{{H}}v_{i}\sigma \left(w_{i}^{T}x+b_{i}\right),
\qquad
| g( x\g\theta ) - f ( x ) | < \delta
\quad
\text{for all }
x\in [0,1]^{m}.
\end{equation*}
The implicit model defined by the collection of functions $g$
and same noise distributions universally approximates the true causal model.
\end{theorem*}

(This directly follows from the approximator theorem of, e.g.,
\citet{cybenko1989approximation}.)
A key aspect is that implicit causal models are not only universal
approximators, but that we can
use fast algorithms for their Bayesian inference (to calculate
\Cref{eq:posterior}).
In particular, variational methods both scale to massive
data and provide accurate posterior approximations
(\Cref{sub:inference}). This lets us obtain good performance in practice
with finite-sized neural networks;
\Cref{sec:results} describes such experiments.

\vspace{-1ex}
\section{Implicit Causal Models with Latent Confounders}
\label{sec:confounders}
\vspace{-1ex}

We described implicit causal models, a rich class of models that can
capture arbitrary causal relations.
For simplicity, we assumed that the global structure is observed; this
enables standard inference methods.
We now consider the typical setting when it is unobserved.

\subsection{Causal Inference with a Latent Confounder}
\label{sub:sharing}

\begin{figure}[tb]
\begin{subfigure}{0.5\columnwidth}
  \centering
  \begin{tikzpicture}

  \factor[minimum size=0.3cm] {x} {} {} {};
  \factor[right=1.1cm of x, minimum size=0.3cm] {y} {} {} {};

  \factor[above=1.1cm of x, xshift=0.75cm, minimum size=0.3cm] {z} {} {} {};

  \node (epsx) [minimum size=0.0cm,left=0.75cm of x] {};
  \node (epsxa) [minimum size=0.0cm,above=0.12cm of epsx.center] {};
  \node (epsxb) [minimum size=0.0cm,below=0.02cm of epsx.center, xshift=-0.2cm] {};
  \node (epsxc) [minimum size=0.0cm,below=0.02cm of epsx.center, xshift=0.2cm] {};
  \draw[] (epsxa.center)--(epsxb.center)--(epsxc.center)--(epsxa.center);

  \node (epsy) [minimum size=0.0cm,right=0.75cm of y] {};
  \node (epsya) [minimum size=0.0cm,above=0.12cm of epsy.center] {};
  \node (epsyb) [minimum size=0.0cm,below=0.02cm of epsy.center, xshift=-0.2cm] {};
  \node (epsyc) [minimum size=0.0cm,below=0.02cm of epsy.center, xshift=0.2cm] {};
  \draw[] (epsya.center)--(epsyb.center)--(epsyc.center)--(epsya.center);

  \node (epsz) [minimum size=0.0cm,left=0.75cm of z] {};
  \node (epsza) [minimum size=0.0cm,above=0.12cm of epsz.center] {};
  \node (epszb) [minimum size=0.0cm,below=0.02cm of epsz.center, xshift=-0.2cm] {};
  \node (epszc) [minimum size=0.0cm,below=0.02cm of epsz.center, xshift=0.2cm] {};
  \draw[] (epsza.center)--(epszb.center)--(epszc.center)--(epsza.center);

  \node[right=0.05cm of z] (zlabel) {$z_n$};
  \node[below=0.05cm of x] (xlabel) {$x_{nm}$};
  \node[below=0.05cm of y] (ylabel) {$y_n$};

  \edge{z}{x};
  \edge{z}{y};
  \edge{x}{y};

  \edge{epsx}{x};
  \edge{epsy}{y};
  \edge{epsz}{z};

  \plate[inner sep=0.1cm,xshift=-0.1cm,yshift=0.15cm,
    label={[xshift=15pt,yshift=15pt]south west:$M$}] {plate1} {
    (x)(xlabel)
    (epsx)
  } {};
  \plate[inner sep=0.2cm,
    label={[xshift=-15pt,yshift=15pt]south east:$N$}] {plate2} {
    (z)(zlabel)
    (x)(xlabel)
    (y)(ylabel)
    (epsx)
    (epsy)
    (epsz)
  } {};

\end{tikzpicture}
\end{subfigure}
\begin{subfigure}{0.5\columnwidth}
  \centering
  \begin{tikzpicture}

  \factor[minimum size=0.3cm] {x} {} {} {};
  \factor[right=1.1cm of x, minimum size=0.3cm] {y} {} {} {};

  \node[above=of x, xshift=-1.9cm, latent, minimum size=0.5cm] (phi)    {};
  \node[above=of y, xshift=1.9cm, latent, minimum size=0.5cm] (theta)    {};
  \node[left=1.5cm of x, latent, minimum size=0.5cm] (w)    {};
  \factor[above=1.1cm of x, xshift=0.75cm, minimum size=0.3cm] {z} {} {} {};

  \node (epsx) [minimum size=0.0cm,yshift=-0.5cm,left=0.75cm of x] {};
  \node (epsxa) [minimum size=0.0cm,above=0.12cm of epsx.center] {};
  \node (epsxb) [minimum size=0.0cm,below=0.02cm of epsx.center, xshift=-0.2cm] {};
  \node (epsxc) [minimum size=0.0cm,below=0.02cm of epsx.center, xshift=0.2cm] {};
  \draw[] (epsxa.center)--(epsxb.center)--(epsxc.center)--(epsxa.center);

  \node (epsy) [minimum size=0.0cm,right=0.75cm of y] {};
  \node (epsya) [minimum size=0.0cm,above=0.12cm of epsy.center] {};
  \node (epsyb) [minimum size=0.0cm,below=0.02cm of epsy.center, xshift=-0.2cm] {};
  \node (epsyc) [minimum size=0.0cm,below=0.02cm of epsy.center, xshift=0.2cm] {};
  \draw[] (epsya.center)--(epsyb.center)--(epsyc.center)--(epsya.center);

  \node (epsz) [minimum size=0.0cm,left=0.75cm of z] {};
  \node (epsza) [minimum size=0.0cm,above=0.12cm of epsz.center] {};
  \node (epszb) [minimum size=0.0cm,below=0.02cm of epsz.center, xshift=-0.2cm] {};
  \node (epszc) [minimum size=0.0cm,below=0.02cm of epsz.center, xshift=0.2cm] {};
  \draw[] (epsza.center)--(epszb.center)--(epszc.center)--(epsza.center);

  \node[below=0.05cm of w] (wlabel) {$w_m$};
  \node[right=0.05cm of z] (zlabel) {$z_n$};
  \node[left=0.05cm of phi] (philabel) {$\phi$};
  \node[right=0.05cm of theta] (thetalabel) {$\theta$};
  \node[below=0.05cm of x] (xlabel) {$x_{nm}$};
  \node[below=0.05cm of y] (ylabel) {$y_n$};

  \edge{w}{x};
  \edge{z}{x};
  \edge{z}{y};
  \edge{phi}{x};
  \edge{theta}{y};
  \edge{x}{y};

  \edge{epsx}{x};
  \edge{epsy}{y};
  \edge{epsz}{z};

  \plate[inner sep=0.1cm,xshift=-0.15cm,yshift=0.15cm,
    label={[xshift=15pt,yshift=15pt]south west:$M$}] {plate1} {
    (x)(xlabel)
    (w)(wlabel)
    (epsx)
  } {};
  \plate[inner sep=0.2cm,
    label={[xshift=-15pt,yshift=15pt]south east:$N$}] {plate2} {
    (z)(zlabel)
    (x)(xlabel)
    (y)(ylabel)
    (epsx)
    (epsy)
    (epsz)
  } {};

\end{tikzpicture}
\end{subfigure}%
\caption{\textbf{(left)}
Causal graph for \gls{GWAS}. The
population structure of \glspl{SNP} for each individual ($z_n$) confounds inference
of how each \gls{SNP} ($x_{nm}$) causes a trait of interest ($y_n$).
\textbf{(right)}
Implicit causal model for \gls{GWAS} (described in
\Cref{sub:implicit-gwas}). Its structure is the same as the causal
graph but also places priors over parameters $\phi$ and $\theta$ and
with a latent variable $w_m$ per \gls{SNP}.
}
\label{fig:gwas}
\end{figure}
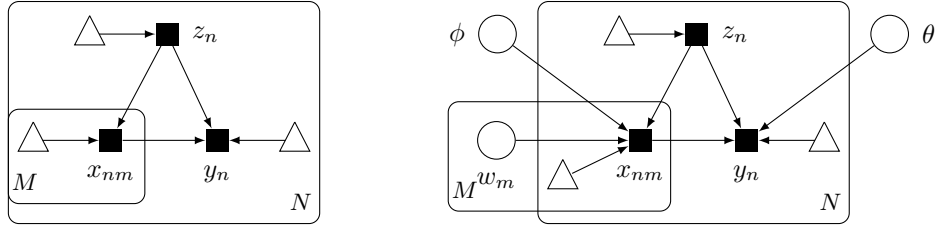

Consider the running example of \glsreset{GWAS}\gls{GWAS} (\Cref{fig:gwas}).  There are $N$
data points (individuals). Each data point consists of an input
vector of length $M$ (measured \glspl{SNP}), $x_n=[x_{n1}, \ldots, x_{nM}]$ and
a scalar outcome $y_n$ (the trait of interest).
Typically, the number of measured \glspl{SNP} $M$ ranges from 100,000
to 1 million and the number of individuals $N$ ranges from 500 to
10,000.

We are interested in how changes to each \gls{SNP} $X_m$ cause
changes to the trait $Y$. Formally, this is
the causal effect
$p(y\g\operatorname{do}(x_{m}), x_{-m})$, which is the probability
of an outcome $y$ given that we force
\gls{SNP} $X_m=x_m$ and consider fixed remaining \glspl{SNP} $x_{-m}$.
Standard inference methods are confounded by the unobserved population
structure of \glspl{SNP} for each individual, as well as the
individual's cryptic relatedness to other samples in the data set.
This confounder is represented as a latent variable $z_n$, which
influences $x_{nm}$ and $y_n$ for each data index $n$; see
\Cref{fig:gwas}.  Because we do not observe the $z_n$'s, the causal
effect $p(y\g\operatorname{do}(x_{m}), x_{-m})$ is unidentifiable
\citep{spirtes1993causation}.

Building on previous \gls{GWAS} methods
\citep{price2006principal,yu2005unified,astle2009population},
we build a model that jointly captures $z_n$'s and the mechanisms for $X_m\to Y$.
Consider the implicit causal model where
for each data point $n=1,\ldots,N$ and
for each \gls{SNP} $m=1,\ldots,M$,
\begin{align}
\begin{split}
z_n &= g_z(\epsilon_{z_n}),
\hspace{8.0em} \epsilon_{z_n}\sim s(\cdot), \\
x_{nm} &= g_{x_m}(\epsilon_{x_{nm}}, z_n\g w_m),
\hspace{3.0em} \epsilon_{x_{nm}}\sim s(\cdot), \\
y_n &= g_y(\epsilon_{y_n}, x_{n,1:M}, z_n\g \theta),
\hspace{2.1em} \epsilon_{y_n}\sim s(\cdot).
\label{eq:model_gwas}
\end{split}
\end{align}
The function $g_z(\cdot)$ for the confounder is fixed. Each
function $g_{x_m}(\cdot\g w_m)$  per \gls{SNP} depends on the
confounder and has parameters
$w_m$. The function $g_y(\cdot\g\theta)$ for
the trait depends on the confounder and all \glspl{SNP},
and it has parameters $\theta$.
We place priors over the parameters
$p(w_m)$ and $p(\theta)$.

\Cref{fig:gwas} (right) visualizes the model.
It is a model over the full causal graph (\Cref{fig:gwas} (left))
and differs from the unconfounded case:
\Cref{eq:posterior} only requires a
model from $X\to Y$, and
the rest of the graph is ``ignorable'' \citep{imbens2015causal}.

To estimate the mechanism $f_y$,
we calculate the posterior of the outcome parameters $\theta$,
\begin{equation}
  \label{eq:shared_posterior}
  p(\theta \g \mbx, \mby)
  = \int p(\mbz\g\mbx,\mby) p(\theta\g\mbx,\mby,\mbz) \d\mbz.
\end{equation}
Note how this accounts for the unobserved confounders: it assumes that
$p(\mbz\g\mbx,\mby)$ accurately reflects the latent structure.
In doing so, we perform
inferences for $p(\theta\g\mbx,\mby,\mbz)$, averaged over posterior
samples from $p(\mbz\g\mbx,\mby)$.

In general,
causal inference with
latent confounders can be dangerous: it uses the data twice (once to
estimate the confounder; another to estimate the mechanism),
and thus it may bias our estimates of each arrow $X_m\to Y$.
Why is this justified? We answer this below.

\begin{proposition}
Assume the causal graph of \Cref{fig:gwas} (left) is correct and that
the true distribution
resides in some configuration of the parameters of the causal model
(\Cref{fig:gwas} (right)).
Then the posterior $p(\theta\g\mbx,\mby)$ provides a consistent estimator
of the causal mechanism $f_y$.
\label{prop:consistency}
\end{proposition}

(See \Cref{appendix:consistency} for the proof.)
\Cref{prop:consistency} rigorizes previous methods in the framework
of probabilistic causal models.
The intuition is that as more \glspl{SNP} arrive (``$M\to\infty$, $N$
fixed''), the posterior concentrates at the true confounders $z_n$, and thus we can
estimate the causal mechanism given each data point's confounder $z_n$.
As more data points arrive (``$N\to\infty$, $M$ fixed''), we can
estimate the causal mechanism given any confounder $z_n$
as there are infinity of them.

\parhead{Connecting to Two-Stage Estimation.}
Existing \gls{GWAS} methods
adjust for latent population structure
using two stages \citep{astle2009population}: first,
estimate the confounders $z_{1:N}$; second, infer the outcome parameters $\theta$
given the data set and the estimate of the confounders.
To incorporate uncertainty, a Bayesian version would not use a point
estimate of $z_{1:N}$ but the full posterior $p(z_{1:N}\g\mbx,\mby)$;
then it would infer $\theta$ given posterior samples of $z_{1:N}$.
Following \Cref{eq:shared_posterior}, this is the same as joint
posterior inference.  Thus the two stage approach is justified as a
Bayesian approximation that uses a point estimate of the posterior.

\subsection{Implicit Causal Model with a Latent Confounder}
\label{sub:implicit-gwas}

Above, we outlined how to specify an implicit causal model for
\gls{GWAS}. We now
specify in detail the functions and priors for
the confounders $z_n$, the \glspl{SNP} $x_{nm}$, and the
traits $y_n$ (\Cref{eq:model_gwas}).
\Cref{fig:gwas} (right) visualizes the model we describe below.
\Cref{appendix:implicit} provides an example implementation in the Edward
probabilistic programming language \citep{tran2016edward}.

\parhead{Generative Process of Confounders $z_n$.}
We use standard normal noise and set the confounder function
$g_z(\cdot)$ to the identity. This implies the distribution of
confounders $p(z_n)$ is standard normal.  Their dimension
$z_n\in\mathbb{R}^K$ is a hyperparameter. The dimension $K$ should be
set to the highest value such that the latent space most closely
approximates the true population structure but smaller than the total
number of \glspl{SNP} to avoid overfitting.

\parhead{Generative Process of \glspl{SNP} $x_{nm}$.}
Designing nonlinear processes that return matrices is an ongoing research direction (e.g.,
\citet{lawrence2005probabilistic,lloyd2012random}).
To design one for \gls{GWAS} (the \gls{SNP} matrix), we build on an implicit modeling
formulation of factor analysis; it has been
successful in \gls{GWAS} applications
\citep{price2006principal,song2015testing}. Let each
\gls{SNP} be encoded as a $0$, $1$, or $2$ to denote the three
possible genotypes. This is unphased data, where 0 indicates two
major alleles; 1 indicates one major and one minor allele; and 2
indicates two minor alleles. Set
\begin{align*}
\operatorname{logit} \pi_{nm} &= z_n^\top w_m
,
\hspace{2.0em}
x_{nm} =
\mathbb{I}[\epsilon_1 > \pi_{nm}] +
\mathbb{I}[\epsilon_2 > \pi_{nm}]
,
\hspace{2.0em} \epsilon_1,\epsilon_2\sim \operatorname{Uniform}(0, 1).
\end{align*}
This defines a $\operatorname{Binomial}(2, \pi_{nm})$
distribution on $x_{nm}$.
Analogous to generalized linear models,
the Binomial's logit probability is linear with respect to $z_n$.
We then sum up two Bernoulli trials: they are represented as indicator
functions of whether a uniform sample is greater than the probability.
(The uniform noises are newly drawn for each index $n$ and
$m$.)

Assuming a standard normal prior on the variables $w_m$,
this generative process is equivalent to logistic factor analysis.
The variables $w_m$ act as ``principal components,'' embedding the
$M$-many \glspl{SNP} within a subspace of lower dimension $K$.

Logistic factor analysis makes strong assumptions: linear dependence
on the confounder and that one parameter per dimension has sufficient
representational capacity.  We relax these assumptions using a neural
network over concatenated inputs,
\begin{equation*}
\operatorname{logit} \pi_{nm} = \operatorname{NN}([z_n, w_m]\g \phi).
\end{equation*}
Similar to the above, the variables $w_m$ serve as principal
components.
The neural network takes an input of dimension $2K$ and
outputs a scalar real value;
its weights and biases $\phi$ are shared across \glspl{SNP} $m$
and individuals $n$. This enables learning of
nonlinear interactions between $z_n$ and $w_m$, preserves the model's conditional independence
assumptions, and avoids the
complexity of a neural net that outputs the full $N\times M$ matrix.
We place a standard normal prior over $\phi$.

\parhead{Generative Process of Traits $y_{n}$.}
To specify the traits, we build on an implicit modeling formulation of
linear regression. It is the mainstay tool in \gls{GWAS} applications
\citep{price2006principal,song2015testing}.  Formally, for real-valued
$y\in\mathbb{R}$, we model each observed trait as
\begin{equation*}
y_n = [x_{n,1:M}, z_n]^\top \theta + \epsilon_n,
\hspace{3.0em} \epsilon_n \sim \operatorname{Normal}(0, 1),
\end{equation*}
This process assumes linear dependence on \glspl{SNP},
no gene-gene and gene-population interaction,
and additive noise.
We generalize this model using a neural
network over the same inputs,
\begin{equation*}
y_n = \operatorname{NN}([x_{n,1:M}, z_n, \epsilon]\g \theta),
\hspace{3.0em} \epsilon_n \sim \operatorname{Normal}(0, 1).
\end{equation*}
The neural net takes an input of dimension $M + K+ 1$ and outputs a
scalar real value; for categorical outcomes, the
output is discretized over equally spaced cutpoints.
We also place a group Lasso prior on weights connecting
a \gls{SNP} to a hidden layer. This encourages sparse inputs: we
suspect few \glspl{SNP} affect the trait \citep{yuan2006model}.
We use standard normal for other weights and biases.

\vspace{-1.0ex}
\section{Likelihood-Free Variational Inference}
\label{sub:inference}
\vspace{-0.5ex}

We described a rich causal model for how
\glspl{SNP} cause traits and that can adjust for latent
population-confounders.
Given \gls{GWAS} data,
we aim to infer the posterior of outcome parameters $\theta$
(\Cref{eq:shared_posterior}).
Calculating this posterior
reduces to calculating the joint posterior of confounders $z_n$,
\gls{SNP} parameters $w_m$ and $\phi$, and trait parameters $\theta$,
\begin{align*}
  p(z_{1:N}, w_{1:M}, \phi, \theta \g \mbx, \mby)
  \propto p(\phi) p(\theta)
  \prod_{n=1}^N
  \Big[
  p(z_n)
  p(y_n \g x_{n,1:M}, z_n, \theta)
  \prod_{m=1}^M
  p(w_m)
  p(x_{nm}\g z_n, w_m,\phi)
  \Big]
  .
\end{align*}
This means we can use typical inference
algorithms on the joint posterior. We then collapse
variables to obtain the marginal posterior of $\theta$. (For
Monte Carlo methods, we drop the auxiliary samples; for
variational methods, it is given if the variational family follows the
posterior's factorization.)

One difficulty is that with implicit models, evaluating the density is
intractable: it requires integrating over a nonlinear function with
respect to a high-dimensional noise (\Cref{eq:implicit}).
Thus we require likelihood-free
methods, which assume that one can only sample from the model's
likelihood \citep{marin2012approximate,tran2017deep}.
Here we apply \gls{LFVI}, which we scale to billions of genetic
measurements \citep{tran2017deep}.

As with all variational methods,
\gls{LFVI} posits a family of distributions over the latent variables
and then optimizes to find the member closest to the posterior.
For the variational family, we specify normal distributions with
diagonal covariance for the \gls{SNP} components $w_m$ and confounder
$z_n$,
\begin{align*}
q(w_m) =
\operatorname{Normal}(w_{m}; \mu_{w_m}, \sigma_{w_m} I),
\hspace{2.5em}
q(z_n) =
\operatorname{Normal}(z_{n}; \mu_{z_n}, \sigma_{z_n} I).
\end{align*}
We specify a point mass for the variational family on both
neural network parameters $\phi$ and $\theta$. (This is equivalent to
point estimation in a variational EM setting.)

For \gls{LFVI} to scale to massive \gls{GWAS} data,
we use stochastic optimization by subsampling \glspl{SNP}
\citep{gopalan2016scaling}.
At a high level, the algorithm proceeds in two stages.
In the first stage, \gls{LFVI} cycles through the following steps:
\vspace{-1ex}
\begin{enumerate}
\item
Sample \gls{SNP} location $m$ and collect the observations at that
location from all individuals.
\vspace{-0.5ex}
\item
Use the observations and current estimate of the confounders
$z_{1:N}$ to
update the $m^{th}$ \gls{SNP} component $w_m$ and \gls{SNP} neural
network parameters $\phi$.
\vspace{-0.5ex}
\item Use the observations and current estimate of
\gls{SNP} components
$w_{1:M}$ to update the
confounders
$z_{1:N}$.
\end{enumerate}
\vspace{-1ex}
This first stage infers the posterior distribution of
confounders $z_n$ and \glspl{SNP} parameters $w_m$ and $\phi$.
Each step's computation is independent of the number of \glspl{SNP},
allowing us to scale to millions of genetic factors. In experiments,
the algorithm converges while scanning over the full set of \glspl{SNP}
only once or twice.

In the second stage, we infer the posterior of outcome parameters
$\theta$ given the inferred confounders from the first stage.
\Cref{appendix:inference} describes the algorithm in more detail;
it expands on the \gls{LFVI} implementation in Edward \citep{tran2016edward}.

\section{Empirical Study}
\label{sec:results}

We described implicit causal models, how to adjust
for latent population-based confounders, and how to perform scalable
variational inference.
In general, validating causal inferences on observational data is not
possible \citep{pearl2000causality}.
Therefore to validate our work, we perform an extensive simulation
study on 100,000 \glspl{SNP}, 940 to 5,000 individuals, and
across 100 replications of 11 settings.
The study indicates that our model is significantly more robust to
spurious associations,
with a state-of-the-art gain of 15-45.3\% in accuracy.
We also apply our model to a real-world \gls{GWAS} of Northern
Finland Birth Cohorts;
our model indeed captures real causal relationships---identifying
similar \glspl{SNP} as previous state of the art.

We compare against three methods that are currently state of the art:
PCA with linear regression \citep{price2006principal} (``PCA''); a
linear mixed model (with the EMMAX software) \citep{kang2010variance}
(``LMM''); and logistic factor analysis with inverse regression
\citep{song2015testing} (``GCAT'').  In all experiments, we use Adam
with a initial step-size of $0.005$, initialize neural network
parameters uniformly with He variance scaling \citep{he2015delving},
and specify the neural networks for traits and \gls{SNP}s as fully
connected with two hidden layers, ReLU activation, and batch
normalization (hidden layer sizes described below). For the trait
model's neural network, we found that including latent variables as
input to the final output layer improves information flow in the
network.

\subsection{Simulation Study: Robustness to Spurious Associations}
\label{sub:simulation}

\begin{figure}[tb]
\centering
\begin{tabular}{lllllll}
\toprule
Trait & \textbf{ICM} & PCA
\textcolor{MidnightBlue}{(Price+06)}
& LMM
\textcolor{MidnightBlue}{(Kang+10)}
& GCAT
\textcolor{MidnightBlue}{(Song+10)}
\\
\midrule
HapMap             & \textbf{99.2} & 34.8 & 30.7 & \textbf{99.2} \\
TGP                & \textbf{85.6} & 2.7 & 43.3 & 70.3 \\
HGDP               & \textbf{91.8} & 6.8 & 40.2 & 72.3 \\
PSD ($a=1$)        & \textbf{97.0} & 80.4 & 92.3 & 95.3 \\
PSD ($a=0.5$)      & \textbf{94.3} & 79.5 & 90.1 & 93.6 \\
PSD ($a=0.1$)      & \textbf{92.2} & 38.1 & 38.6 & 90.4 \\
PSD ($a=0.01$)     & \textbf{92.7} & 24.2 & 35.1 & 90.7 \\
Spatial ($a=1$)    & \textbf{90.9} & 56.4 & 60.0 & 75.2 \\
Spatial ($a=0.5$)  & \textbf{86.2} & 50.5 & 46.6 & 72.5 \\
Spatial ($a=0.1$)  & \textbf{80.9} & 2.4 & 26.6 & 35.6 \\
Spatial ($a=0.01$) & \textbf{75.5} & 1.8 & 15.3 & 30.2 \\
\bottomrule
\end{tabular}
\captionof{table}{Precision accuracy over an extensive set of
configurations and methods; we average over 100 simulations
for a grand total of 4,400 fitted models.
The setting $a$ in PSD and Spatial determines the amount of sparsity
in the latent population structure: lower $a$ means higher sparsity.
ICM is significantly more robust to spurious associations,
outperforming other methods by up to 45.3\%.
}
\label{table:simulation-study}
\end{figure}

We analyze 11 simulation configurations, where each configuration
uses 100,000 SNPs and 940 to 5,000 individuals.
We simulate 100 \gls{GWAS} data sets per configuration for a grand
total of 4,400 fitted models (4 methods of comparison).
Each configuration employs
a true model to generate the \glspl{SNP} and traits
based on real genomic data.
Following \citet{hao2016probabilistic}, we use
the Balding-Nichols model based on the HapMap dataset
\citep{balding1995method,gibbs2003international};
PCA based on the 1000 Genomes Project (TGP) \citep{thousand2010map};
PCA based on the Human Genome Diversity project (HGDP)
\citep{rosenberg2002genetic};
four variations of the Pritchard-Stephens-Donelly model (PSD) based on HGDP
\citep{pritchard2000inference};
and four variations of a configuration where population structure
is determined by a latent spatial position of individuals.
Only 10 of the 100,000 \glspl{SNP} are set to be causal.
\Cref{appendix:simulation} provides more detail.

\Cref{table:simulation-study} displays the precision for predicting
causal factors across methods.
When failing to account for population structure, ``spurious
associations'' occur between genetic markers and the trait of
interest, despite the fact that there is no biological connection.
Precision is the fraction of the
number of true positives over the number of true and false positives.
This measures a method's robustness to spurious associations:
higher precision means fewer false positives and thus more robustness.

\Cref{table:simulation-study} shows that our method achieves state of
the art across all configurations.  Our method especially dominates in
difficult tasks with sparse (small $a$), spatial (Spatial), and/or
mixed membership structure (PSD): there is over a 15\% margin in
difference to the second best in general, and up to a 45.3\% margin on
the Spatial ($a=0.01$) configuration.  For simpler configurations,
such as a mixture model (HapMap), our method has comparable
performance.

\subsection{Northern Finland Birth Cohort Data}
\label{sub:northern}

\begin{table*}[tb]
\centering
\begin{tabular}{lllllll}
\toprule
Trait & \textbf{ICM} & GCAT & LMM & PCA & Uncorrected \\
\midrule
Body mass index & 0 & 0 & 0 & 0 & 0 \\
C-reactive protein & 2 & 2 & 2 & 2 & 2 \\
Diastolic blood pressure & 0 & 0 & 0 & 0 & 0 \\
Glucose levels & 3 & 3 & 2 & 2 & 2  \\
HDL cholesterol levels & 4 & 4 & 4 & 2 & 4  \\
Height & 1 & 1 & 0 & 0 & 0  \\
Insulin levels & 0 & 0 & 0 & 0 & 0  \\
LDL cholesterol levels & 3 & 4 & 3 & 3 & 3  \\
Systolic blood pressure & 0 & 0 & 0 & 0 & 0  \\
Triglyceride levels & 2 & 2 & 3 & 2 & 2  \\
\bottomrule
\end{tabular}
\caption{Number of significant loci at genome-wide significance $(p < 7.2\times 10^{-8})$ for each of the ten traits from NFBC data.
The counts for GCAT are obtained from
\citet[Table 1]{song2015testing};
counts for LMM, PCA, and uncorrected are obtained from
\citet[Table 2]{kang2010variance}.
The implicit causal model (\textbf{ICM}) captures causal relationships comparable to previously work.
}
\label{table:real-data}
\end{table*}
We analyze a real-world \gls{GWAS} of Northern Finland
Birth Cohorts \citep{sabatti2009genome}, which measure several
metabolic traits and height and which contain 324,160 \glspl{SNP} and
5,027 individuals.
We separately fitted 10 implicit causal models, each of which models
the effect of \glspl{SNP} on one of ten traits.
To specify the implicit causal models,
we set the latent dimension of confounders to be 6 (following
\citet{song2015testing}).
We use 512 units in both hidden layers of the
\gls{SNP} neural network and use 32 and 256 units for the
trait neural network's first and second hidden layers respectively.
\Cref{appendix:northern} provides more detail.

\Cref{table:real-data} compares the number of identified causal
\glspl{SNP} across methods, with an additional
``uncorrected'' baseline, which does not adjust for any latent population structure.
Each method is performed with a subsequent correction as measured by
the genomic control inflation factor \citep{sabatti2009genome}.  Our
models identify similar causal \glspl{SNP} as previous methods.
Interestingly, our model tends to agree with \citet{song2015testing},
identifying a total of 15 significant loci (\citet{song2015testing}
identified 16; others identified 11-14 loci).  This makes sense
intuitively, as \citet{song2015testing} uses logistic factor analysis
which, compared to all methods, most resembles our model.

\section{Discussion}

We described implicit causal models, a rich class of models that can
capture high-dimensional, nonlinear causal relationships.  With
\acrlong{GWAS}, implicit causal models generalize previous successful
methods to capture important nonlinearities, such as gene-gene and
gene-population interaction. In addition, we described an implicit
causal model that
adjusts for confounders by sharing strength across examples.  Our
model achieves state-of-the-art accuracy, significantly outperforming
existing genetics methods by 15-45.3\%.

While we focused on \gls{GWAS} applications in this paper,
we also believe implicit causal models have significant potential in other
sciences: for example, to design new dynamical theories in high energy
physics; and to accurately model structural equations of discrete
choices in economics. We're excited about applications to these new
domains, leveraging modern probabilistic modeling and causality to
drive new scientific understanding.

\paragraph{Acknowledgements.}
DT is supported by a Google Ph.D. Fellowship in Machine Learning and
an Adobe Research Fellowship.
DMB is supported by NSF IIS-0745520, IIS-1247664, IIS-1009542, ONR N00014-11-1-0651, DARPA FA8750-14-2-0009, N66001-15-C-4032 , Facebook, Adobe, Amazon, and the John Templeton Foundation.

\bibliographystyle{iclr2018_conference}
\bibliography{bib}
\clearpage
\appendix

\section{Consistency}
\label{appendix:consistency}

Consider the simplest setting in \Cref{sec:method}, where the causal graph
is as shown with a global confounder with finite dimension and where
we observe the data set $\{(x_n, y_n)\}$. Assume the specified model
family over the causal graph includes the true data generating
process.

First consider
consider an atomic intevention $\operatorname{do}(X=x)$ and let
$\beta^*$ be the true structural value that generated our
observations. The probability of a new outcome given
the intervention and global structure is
\begin{align*}
  p^*(y\g\operatorname{do}(X=x), \beta^*))
  &= p^*(y\g x, \beta^*).
\end{align*}
This follows from the backdoor criterion on the empty set.
By Bernstein
von-Mises, the posterior for our model $p(\beta\g\mbx,\mby)$
concentrates at $\beta^*$. Thus, similarly, our posterior for $\theta$ given
$\beta^*$ concentrates to the true functional mechanism
$f_y$. This implies we have a consistent estimate of
$p^*(y\g\operatorname{do}(\cdot), \beta^*)$.

This simple proof rigorizes the ideas behind learning and fixing
population structure, a common heuristic in \gls{GWAS} methods.
Moreover, it lets us understand how to extend them to more complex
latent variable models of the confounder and also provide uncertainty
estimates of the latent structure become important as we apply these
methods to finite data in practice.

\section{Implicit Causal Model in Edward}
\label{appendix:implicit}

We provide an example of an implicit causal model written in the
Edward language below. It writes neural net weights and biases as
model parameters. It does not include priors on the weights or biases;
we add these as penalties to the objective function during training.

\begin{lstlisting}[language=Python]
import edward as ed
import tensorflow as tf
from edward.models import Binomial, Normal

N = 5000  # number of individuals
M = 100000  # number of SNPs
K = 25  # latent dimension

def snp_neural_network(z, w):
  z_tile = tf.tile(tf.reshape(z, [N, 1, K]), [1, M, 1])
  w_tile = tf.tile(tf.reshape(w, [1, M, K]), [N, 1, 1])
  h = tf.concat([z_tile, w_tile], 2)
  h = tf.layers.dense(h, 512, activation=tf.nn.relu)
  h = tf.layers.dense(h, 512, activation=tf.nn.relu)
  h = tf.layers.dense(h, 1, activation=None)
  return tf.reshape(h, [N, M])

def trait_neural_network(z, x):
  eps = Normal(loc=0.0, scale=1.0, sample_shape=[N, 1])
  h = tf.concat([z, x, eps], 1)
  h = tf.layers.dense(h, 32, activation=tf.nn.relu)
  h = tf.layers.dense(h, 256, activation=tf.nn.relu)
  h = tf.concat([z, h], 1)  # include connection to z for output layer
  h = tf.layers.dense(h, 1, activation=None)
  return tf.reshape(h, [N])

z = Normal(loc=0.0, scale=1.0, sample_shape=[N, K])
w = Normal(loc=0.0, scale=1.0, sample_shape=[M, K])
logits = snp_neural_network(z, w)
x = Binomial(total_count=2.0, logits=logits)

# Note: Sometimes it's preferable to have neural net parameterize a
# distribution for tractable density such as for binary-valued traits.
# To do this, specify, e.g., `y = Bernoulli(logits=trait_neural_network(z, x))`
y = trait_neural_network(z, x)
\end{lstlisting}

\section{Likelihood-Free Variational Inference for GWAS}
\label{appendix:inference}

\subsection{Variational Objective}

The log-likelihood per-individual and per-\gls{SNP} is
\begin{align*}
\log p(x_{nm},y_n\g w_m,z_n,\theta,\phi)
=
\log p(y_n\g x_{n,1:M}, z_n, \theta)
+ \log p(x_{nm}\g w_m, z_n, \phi).
\end{align*}
There are local priors $p(w_m)$, $p(z_n)$ and global
priors $p(\theta)$, $p(\phi)$.
The posterior factorizes as
\begin{align*}
&
p(z_{1:N}, w_{1:M}, \phi, \theta \g \mbx, \mby)
\\
&\quad
= p(\phi\g\mbx) p(\theta\g\mbx,\mby)
  \prod_{m=1}^M p(w_m\g x_{1:N,M},\phi)
  \prod_{n=1}^N p(z_n\g x_{n,1:M}, y_n,\theta,\phi,w_{1:M}).
\end{align*}
Let the variational family follow the posterior's factorization above.
For notational convenience, we drop the data dependence in $q$,
$q(\phi)=q(\phi\g\mbx)$,
$q(\theta)=q(\theta\g\mbx,\mby)$
$q(w_m\g\phi)=q(w_m\g x_{1:N,m},\phi)$,
$q(z_n\g\theta,\phi,w_{1:M})=q(z_n\g x_{n,1:M},y_n,\theta,\phi,w_{1:M})$.
We write the evidence lower bound and
decompose the model's log joint density and the variational density,
\begin{align*}
\mathcal{L}
&=
\sum_{m=1}^M
\sum_{n=1}^N
\mathbb{E}_{q(\phi)q(w_m\g\phi)q(\theta)q(z_n\g\theta,\phi,w_{1:M})}\Big[
\log p(x_{nm}\g w_m, z_n, \phi)
\Big] +
\\
&\quad
\sum_{n=1}^N
\mathbb{E}_{q(\phi)q(w_{1:M}\g\phi)q(\theta)q(z_n\g\theta,\phi,w_{1:M})}\Big[
\log p(y_n\g x_{n,1:M}, z_n, \theta)
\Big] +
\\
&\quad
\sum_{m=1}^M
\mathbb{E}_{q(\phi)q(w_m\g\phi)}\Big[
\log p(w_m)
- \log q(w_m\g\phi)
\Big] +
\\
&\quad
\sum_{n=1}^N
\mathbb{E}_{q(\phi)q(w_{1:M}\g\phi)q(\theta)q(z_n\g\theta,\phi,w_{1:M})}\Big[
\log p(z_n)
- \log q(z_n\g\theta,\phi,w_{1:M})
\Big] +
\\
&\quad
\mathbb{E}_{q(\phi)}[\log p(\phi) - \log q(\phi)] +
\mathbb{E}_{q(\theta)}[\log p(\theta) - \log q(\theta)]
.
\end{align*}
Assume that the variational family for $z_n$ and $w_m$
are independent of other variables, $q(z_n)$
and $q(w_m)$. Also assume delta point masses for
$q(\theta)=\mathbb{I}[\theta - \theta']$ parameterized by $\theta'$ and
$q(\phi)=\mathbb{I}[\phi - \phi']$ parameterized by $\phi'$. This
simplifies the objective, reducing to
\begin{align*}
\mathcal{L}
&=
\sum_{m=1}^M
\sum_{n=1}^N
\mathbb{E}_{q(w_m)q(z_n)}\Big[
\log p(x_{nm}\g w_m, z_n, \phi')
\Big] +
\sum_{n=1}^N
\mathbb{E}_{q(z_n)}\Big[
\log p(y_n\g x_{n,1:M}, z_n, \theta')
\Big] +
\\
&\quad
\sum_{m=1}^M
\mathbb{E}_{q(w_m)}\Big[
\log p(w_m)
- \log q(w_m)
\Big] +
\sum_{n=1}^N
\mathbb{E}_{q(z_n)}\Big[
\log p(z_n)
- \log q(z_n)
\Big] +
\\
&\quad
\log p(\phi') +
\log p(\theta')
.
\end{align*}
Each expectation can be unbiasedly estimated with Monte Carlo.
For gradient-based optimization, we use reparameterization gradients
\citep{rezende2014stochastic}. We describe them next.

\subsection{First Stage: Learning the Confounder}

We provide details for the gradients.
Let $\lambda_{w_m}$ parameterize $q(w_m; \lambda_{w_m})$ and
$\lambda_{z_n}$ parameterize $q(z_n; \lambda_{z_n})$.
We are interested in training the parameters
$\lambda_{w_m}, \lambda_{z_n}, \theta', \phi'$.
Subsample a \gls{SNP} location $m\in\{1,\ldots,M\}$.
Draw a sample $z_n'\sim q(z_n; \lambda_{z_n})$ for $n=1,\ldots,N$
and $w_m'\sim q(w_m; \lambda_{w_m})$ for $m=1,\ldots,M$, where the samples are
reparameterizable
(see \citet{rezende2014stochastic} for details).

The gradient with respect to parameters
$\lambda_{z_n}$
is unbiasedly estimated by
\begin{align*}
\nabla_{\lambda_{z_n}}
\mathcal{L}
&\approx
\nabla_{\lambda_{z_n}}
\Big[
\log p(x_{nm}\g w_m', z_n', \phi')
+
\log p(y_n\g x_{n,1:M}, z_n', \theta')
+
\log p(z_n')
-
\log q(z_n')
\Big]
.
\end{align*}
This gradient scales linearly with the number of
\glspl{SNP} $M$. This is undesirable as the number of \glspl{SNP}
ranges from the hundreds of thousands to millions.
We prevent this scaling by observing that for large $M$, the
information in $x_{n,1:M}$ will influence the posterior far more than the
single scalar $y_n$.
In math,
$p(z_n\g x_{n,1:M}, y_n, \theta, \phi, w_{1:M})\approx
p(z_n\g x_{n,1:M}, \theta, \phi, w_{1:M})$.
This is a tacit assumption in all \gls{GWAS} methods that adjust for the
confounder \citep{yu2005unified,price2006principal,astle2009population,kang2010variance}.

The gradient with respect to parameters $\lambda_{z_n}$ simplifies to
\begin{align*}
\nabla_{\lambda_{z_n}}
\mathcal{L}
&\approx
\nabla_{\lambda_{z_n}}
\Big[
\log p(x_{nm}\g w_m', z_n', \phi')
+
\log p(z_n')
-
\log q(z_n')
\Big],
\end{align*}
which scales to massive numbers of \glspl{SNP}.

The gradients with respect to parameters
$\lambda_{w_m}$ and $\phi'$
are
unbiasedly estimated by
\begin{align*}
\nabla_{\lambda_{w_m}}
\mathcal{L}
&\approx
\sum_{n=1}^N
\Big[
\nabla_{\lambda_{w_m}}
\log p(x_{nm}\g w_m', z_n', \phi')
\Big] +
\nabla_{\lambda_{w_m}}
\log p(w_m')
-
\nabla_{\lambda_{w_m}}
\log q(w_m')
,
\\
\nabla_{\phi'}
\mathcal{L}
&\approx
\sum_{n=1}^N
\nabla_{\phi'}
\log p(x_{nm}\g w_m', z_n', \phi')
.
\end{align*}
Note how none of this depends on the trait $y$ or trait parameters $\theta$. We can
thus perform inference to first approximate the posterior
$p(z_{1:N}, w_{1:M}, \phi\g \mbx,\mby)$.
In a second stage, we can then perform inference to approximate the posterior
$p(\theta\g z_{1:N}, w_{1:M}, \mbx,\mby)$.
The computational savings is significant not only within task but
across tasks:
when modelling many traits of interest (for example,
\Cref{sub:northern}), inference over the \gls{SNP} confounders only
needs to be done once and can be re-used.
We perform stochastic gradient ascent using these gradients to maximize the
variational objective.

\subsection{Second Stage: Learning the Trait}

Above we described the first stage of an algorithm which performs stochastic gradient ascent to optimize parameters so that
$q(z_n)q(w_{1:M})q(\theta')\approx p(z_{1:N}, w_{1:M}, \phi\g \mbx,\mby)$.
Given these parameters, we are interested in training
$\theta'$. Dropping constants with respect to $\theta$ in the
objective, we have
\begin{align*}
\mathcal{L}
&\propto
\sum_{n=1}^N
\mathbb{E}_{q(z_n)}\Big[
\log p(y_n\g x_{n,1:M}, z_n, \theta')
\Big] +
\log p(\theta')
.
\end{align*}
We maximize this objective using stochastic gradients with a single
sample $z_n'\sim q(z_n)$,
\begin{align*}
\nabla_{\theta'}
\mathcal{L}
&\approx
\sum_{n=1}^N
\nabla_{\theta'}
\log p(y_n\g x_{n,1:M}, z_n', \theta')
+
\nabla_{\theta'}
\log p(\theta')
\end{align*}
This corresponds to Monte Carlo EM. Its primary computation per-iteration
is the backward pass of the trait's neural network.
Unlike in the first stage, we do not subsample \glspl{SNP} as the
likelihood depends on all \glspl{SNP}.

\subsection{Second Stage: Handling Likelihood-Free Traits}

In general, the density of $y_n$ is intractable: we exploit its
tractable density if $y_n$ is discrete (it induces a
categorical likelihood); otherwise for real-valued traits, we perform
likelihood-free inference with respect to $y_n$. Following
\gls{LFVI} \citep{tran2017deep},
define $q(y)$ to be the empirical distribution over observed data $\{y_n\}$.
Then subtract it as a constant to the objective, so
\begin{equation*}
\mathcal{L}
\propto
\sum_{n=1}^N
\mathbb{E}_{q(z_n)}\Big[
\log p(y_n\g x_{n,1:M}, z_n, \theta')
-
\log q(y_n)
\Big] +
\log p(\theta').
\end{equation*}
We approximate this log-ratio with a ratio estimator,
$r(y_n, x_{n,1:M}, z_n, \theta'; \lambda_r)$.
It is a function of all inputs in the log-ratio and is parameterized
by $\lambda_r$.

We train the ratio estimator
by minimizing a loss function with respect to
its parameters,
\begin{equation*}
\mathbb{E}_{p(y_n\g x_{n,1:M}, z_n, \theta')}
[ -\log \sigma(r(y_n, x_{n,1:M}, z_n, \theta'; \lambda_r)) ]
+
\mathbb{E}_{q(y_n)}
[ -\log (1 - \sigma(r(y_n, x_{n,1:M}, z_n, \theta'; \lambda_r))) ]
.
\end{equation*}
The global minima of this objective with respect to the ratio
estimator is the desired log-ratio,
\begin{equation*}
r^*(y_n,x_{n,1:M}, z_n,\theta') =
\log p(y_n\g x_{n,1:M}, z_n, \theta')
- \log q(y_n).
\end{equation*}
Unfortunately, the ratio estimator has inputs of many dimensions. In particular,
it has the problematic property of scaling with the number of
\glspl{SNP}, which can be on the order of hundreds of thousands.

We can efficiently parameterize the ratio estimator by studying two
extreme cases with respect to computational efficiency and statistical
efficiency. In one extreme, suppose $y_n$ has a tractable Gaussian
density with mean given by the outcome model's neural network and unit variance
(that is, the neural net
is parameterized to apply a location-shift on the noise input, $y_n=\operatorname{NN}(\cdot)
+ \epsilon_n$).
Up to additive and
multiplicative constants, the optimal log-ratio is
\begin{equation*}
r^*(y_n,x_{n,1:M}, z_n,\theta')
\propto (y_n - \operatorname{NN}([x_{n,1:M}, z_n]\g
\theta'))^2.
\end{equation*}
This implies the ratio estimator must relearn
the neural network's forward pass in order to estimate the optimal log-ratio.
This is computationally redundant and can lead to unstable training.
On another extreme, suppose we parameterize $r$ as
\begin{equation*}
r(y_n,x_{n,1:M}, z_n,\theta'; \lambda_r) =
r(y_n,\operatorname{NN}([x_{n,1:M}, z_n, \epsilon]\g\theta'); \lambda_r).
\end{equation*}
This dramatically reduces $r$'s input dimensions, from hundreds of
thousands to just two.
However, while computationally efficient, this is a poor statistical approximation: there
is only a single dimension to preserve information about
$x_{n,1:M}$, $z_n$, $\epsilon_n$, and $\theta'$ relevant to $y_n$;
this dimension is lossy for even Gaussian densities.

As a middleground, we use
the neural net's first hidden layer as input into the ratio estimator,
\begin{equation*}
r(y_n,x_{n,1:M}, z_n,\theta'; \lambda_r) =
r(y_n, h_n; \lambda_r).
\end{equation*}
This reduces the ratio estimator's inputs to be
the trait $y_n$ and first hidden layer's units $h_n$. This hidden layer has
much fewer dimensions than the raw inputs, such as 32 units (making it computationally
efficient). Moreover, under the data processing inequality
\citep{cover1991elements}, $h_n$ preserves more information relevant to $y_n$ than subsequent layers of the neural network (making it statistically efficient).
For all experiments, we parameterized $r$ with two fully connected
hidden layers with equal number of hidden units.

The gradient with respect to parameters
$\theta$
is
estimated by
\begin{align*}
\nabla_{\theta'}
\mathcal{L}
&\approx
\sum_{n=1}^N
\nabla_{\theta'}
r(y_n, x_{n,1:M}, z_n', \theta'; \lambda_r)
+
\log p(\theta')
.
\end{align*}
This substitutes in the ratio estimator as a proxy to the intractable
likelihood.

We minimize the auxiliary loss function in
order to train the ratio estimator.
Sample $y_n'\sim p(y_n\g x_{n,1:M}, z_n, \theta')$ and
subsample a data point $y_n\sim q(y_n)$.
The gradient is estimated by
\begin{equation*}
\nabla_{\lambda_r}\cdot \approx
\nabla_{\lambda_r}
\Big[
-\log \sigma(r(y_n', h_n; \lambda_r))
-
\log (1 - \sigma(r(y_n, h_n; \lambda_r)))
\Big]
.
\end{equation*}
This corresponds to maximum likelihood, balanced with an
adversarial objective to estimate the likelihood, and is relatively
fast.
We perform stochastic gradient ascent, alternating between these
two sets of gradients.

\section{Simulation Study}
\label{appendix:simulation}

We provide more detail to \Cref{sub:simulation}.
Implicit causal models can not only represent many causal structures
but, more importantly, learn them from data.
To demonstrate this, we simulate data from a comprehensive collection
of popular models in \gls{GWAS} and analyze how well the fitted model
can capture them.
These configurations exactly follow \citet{hao2016probabilistic} with
same hyperparameters, which we describe below.

For each of the 11 simulation configurations, we generate 100
independent data sets.
Each data set consists of a $M\times N$ matrix of genotypes $X$ and
vector of $N$ traits $y$. Each individual $n$ has $M$
\glspl{SNP} and one trait.

\subsection{Genotype Matrix}

To simulate the $M \times N$ matrix of genotypes $X$,
we draw $x_{mn}\sim\operatorname{Binomial}(2,\pi_{mn})$ for $m=1,\ldots,M$
\glspl{SNP} and $n=1,\ldots,N$ individuals.
The probabilities $\pi_{mn}$ can be encoded under a real-valued
$M\times N$ matrix of allele frequencies $F$ where $\pi_{mn} =
[\operatorname{logit}(F)]_{mn}$.

Many models in \gls{GWAS} can be described under the factorization
$F = \Gamma S$,
where $\Gamma$ is a $M\times K$ matrix and $S$ is a $K\times N$ matrix
for a fixed rank $K\le N$.
This includes
principal components analysis \citep{price2006principal},
the Balding-Nichols model \citep{balding1995method},
and the
Pritchard-Stephens-Donnelly model \citep{pritchard2000inference}.
The $M\times K$ matrix $\Gamma$ describes how structure manifests in
the allele frequencies across \glspl{SNP}.
The $K\times N$ matrix $S$ encapsulates the genetic population
structure across individuals.

We describe how we form $\Gamma$ and $S$ for each of the 11 simulation
configurations.

\parhead{Balding-Nichols Model (BN) + HapMap.}
The BN model describes individuals according to a discrete mixture of
ancestral subpopulations \citep{balding1995method}.
The HapMap data set was collected from three discrete
populations \citep{gibbs2003international},
which allows us to populate each row $m$ of $\Gamma$
with three i.i.d. draws from the Balding-Nichols model:
$\Gamma_{mk} \sim \operatorname{BN}(p_m,F_m)$, where $k\in\{1,2,3\}$.
Each $\Gamma_{mk}$ is interpreted to be the allele frequency for
subpopulation $k$ at SNP $m$. The pairs $(p_m, F_m)$ are computed by
randomly selecting a SNP in the HapMap data set, calculating its
observed allele frequency, and estimating its $F_{ST}$ value
using the estimator
of \citet{weir1984estimating}. The columns of $S$ are populated with
indicator vectors such that each individual is assigned to one of the
three subpopulations. The subpopulation assignments are drawn
independently with probabilities $60/210$, $60/210$, and $90/210$,
which reflect the subpopulation proportions in the HapMap data set.
The simulated data has $M=100,000$ \glspl{SNP} and $N=5000$
individuals.

\parhead{1000 Genomes Project (TGP).}
TGP is a project that comprehensively catalogs human genetic variation
by producing complete genome sequences of well over 1000 individuals
of diverse ancestries \citep{thousand2010map}.
To form $\Gamma$, we sample
$\Gamma_{mk} \sim 0.9\operatorname{Uniform}(0, 1/2)$ for
$k=1,2$ and set $\Gamma_{m3} = 0.05$. To form $S$,
we compute the first two principal components of the TGP genotype
matrix after mean centering each \gls{SNP}. We then transform each
principal component to be between $(0,1)$ and set the first two rows
of $S$ to be the transformed principal components. The third row of
$S$ is set to $1$ as an intercept. The simulated data has $M=100,000$
\glspl{SNP} and $N=1500$ individuals (the total number of individuals
in the TGP data set).

\parhead{Human Genome Diversity Project (HGDP).}
HGDP is an international project that has genotyped a large collection
of DNA samples from individuals distributed around the world, aiming
to assess worldwide genetic diversity at the genomic level
\citep{rosenberg2002genetic}.
We followed the same scheme as for TGP above.
The simulated data has $M=100,000$ \glspl{SNP} and $N=940$
individuals (the total number of individuals in the HGDP data set).

\parhead{Pritchard-Stephens-Donnelly (PSD) + HGDP.}
The PSD model describes individuals according to an admixture of
ancestral subpopulations \citep{pritchard2000inference}.
The rows of $\Gamma$ are drawn from three
i.i.d. draws from the Balding-Nichols model:
$\Gamma_{mk} \sim \operatorname{BN}(p_m,F_m)$, where
$k\in\{1, 2, 3\}$.
The pairs $(p_m, F_m)$ are computed by
randomly selecting a SNP in the HGDP data set, calculating its
observed allele frequency, and estimating its $F_{ST}$ value
using the estimator
of \citet{weir1984estimating}.
The estimator requires each individual to be assigned to a
subpopulation, which are made
according to the $K = 5$ subpopulations from the analysis in
\citet{rosenberg2002genetic}.
The columns of $S$ are sampled $(s_{1n},s_{2n},s_{3n}) \sim
\operatorname{Dirichlet}(\mbalpha=(a,a,a))$ for $n = 1,\ldots,N$. We apply
four PSD configurations with hyperparameter settings of $a =
0.01,0.1,0.5,1$.
Varying $a$ from 1 to 0 varies the level of sparsity as
individuals are placed from uniformly to corners of the simplex.
The simulated data has $M=100,000$ \glspl{SNP} and $N=5000$ individuals.

\parhead{Spatial. }
In this setting, we simulate genotypes such that the population
structure relates to the
spatial position of individuals. The matrix $\Gamma$ is populated
by sampling $\Gamma_{mk} \sim 0.9\operatorname{Uniform}(0, 1/2)$ for
$k = 1,2$ and setting $\Gamma_{m3} = 0.05$. The first two rows of $S$
correspond to coordinates for each individual on the unit square and
are set to be independent and identically distributed samples from
$\operatorname{Beta}(a, a)$, while the third row of $S$ is set to
$1$ as an intercept. We apply four spatial configurations with
hyperparameter settings of
$a = 0.01, 0.1, 0.5, 1$.
As with the Dirichlet distribution in the PSD model, varying
$a$ from 1 to 0 varies the level of sparsity as individuals are placed
from uniformly to corners of the unit square.
The simulated data has $M=100,000$ \glspl{SNP} and $N=5000$ individuals.

\subsection{Traits of Interest}

To simulate traits $y$, we simulate from a linear model: for each
individual $n$'s \glspl{SNP} $\{x_{mn}\}$,
\begin{equation*}
y_n = \sum_{m=1}^M\beta_m x_{mn} + \lambda_n + \epsilon_n,
\qquad
\epsilon_n \sim\operatorname{Normal}(0, \sigma_n^2).
\end{equation*}
Each trait is real-valued and is determined by a linear combination of
\glspl{SNP}, a per-individual offset, and heteroskedastic noise.
Below we describe how we set $\{\beta_m\}$, $\{\lambda_n\}$, and
$\{\sigma_n\}$.

Without loss of generality, we set the first 10 \glspl{SNP} to be true
alternative \glspl{SNP} ($\beta_m\neq0$), where
$\beta_m\sim\operatorname{Normal}(0, 0.5)$ for $m=1,2,\ldots,10$;
$\beta_m = 0$ for $m > 10$.
In order to simulate $\lambda_j$ and $\epsilon_j$ so that they are
also influenced by the latent variables $z_1, \ldots, z_n$, we
performed the following:

\begin{enumerate}
\item
Run $K$-means clustering on the columns of $S$ with $K = 3$ using
Euclidean distance. This assigns each individual $j$ to one of three
partitions $S_1,S_2,S_3$ where $S_k \subset {1,2,\ldots,n}$.
\item
Set $\lambda_j =k$ for all $j\in S_k$ for each $k=1,2,3$.
\item
Draw $\tau_1^2,\tau_2^2,\tau_3^2 \sim \operatorname{InverseGamma}(3,1)$.
Set $\sigma_j =\tau_k$ for all $j\in S_k$.
\end{enumerate}

Following \citet{song2015testing},
\begin{leftbar}
This strategy simulates non-genetic effects and random variation that
manifest among $K$ discrete groups over a more continuous population
genetic structure defined by $S$. This is meant to emulate the fact that environment (specifically lifestyle) may partition among individuals in a manner distinct from, but highly related to population structure.
\end{leftbar}

We apply this strategy for each of the 11 configurations where each
involved up to $M=100,000$ \glspl{SNP} and $N=5000$ individuals.
For each configuration, we simulated 100 independent data sets, thus
requiring a total of 1100 fitted models per method of comparison.

\subsection{Other Details}

After fitting each model, we obtain one $p$-value for each \gls{SNP},
which is the probability that the \gls{SNP}'s effect on the trait of
interest is zero.
We fix a $p$-value threshold of $t=0.0025$.
To calculate the number of observed positives, we count the number
of $p$-values for that are less than or equal to $t$.
The true positives are the subset of $p$-values associated with causal \glspl{SNP};
false positives are those associated with null \glspl{SNP}.

Spurious associations occur when $p$-values corresponding to
null \glspl{SNP} are artificially small.
Namely, false positives are spurious associations.
In general, we expect there to be $m_0\times t$ false positives among
the $m_0$ $p$-values corresponding to null \glspl{SNP}; in our
setting, this corresponds to $(100,000 - 10)\cdot0.0025\approx 250$
\glspl{SNP}.
A method properly accounts for structure when the average excess is no
more than this number.

To specify the implicit causal model in our experiments, we set the
latent dimension of confounders equal to 3 or 5.  We use 512 units in
both hidden layers of the
\gls{SNP} neural network and use 32 and 256 units for the
trait neural network's first and second hidden layers respectively.

\section{Northern Finland Birth Cohort Data}
\label{appendix:northern}

We provide more detail to \Cref{sub:northern}.
The data was obtained from the database of Genotypes and Phenotypes
(dbGaP) (\texttt{phs000276.v1.p1}).
We follow the same preprocessing as \citet{song2015testing},
\begin{leftbar}
Individuals were filtered for completeness (maximum 1\% missing
genotypes) and pregnancy. (Pregnant women were excluded because we did
not receive IRB approval for these individuals.) \glspl{SNP} were first
filtered for completeness (maximum 5\% missing genotypes) and minor
allele frequency (minimum 1\% minor allele frequency), then tested for
Hardy-Weinberg equilibrium (p-value < $1/328348$).  The final
dimensions of the genotype matrix are $m = 324,160$ \glspl{SNP} and $n =
5027$ individuals.

A Box-Cox transform was applied to each trait, where the parameter was
chosen such that the values in the median 95\% value of the trait was
as close to the normal distribution as possible. Indicators for sex,
oral contraception, and fasting status were added as adjustment
variables. For glucose, the individual with the minimum value was
removed from the analysis as an extreme outlier.
\end{leftbar}
No additional changes were made to the data.

After fitting each model, we follow the same procedure as in the
simulation study for predicting causal factors. We set the $p$-value threshold
to be the genome-wide threshold of $7.2\times 10^{-8}$ following \citet{kang2010variance}.

\end{document}